\documentclass{article}
\usepackage[letterpaper]{geometry}
\usepackage{spconf,graphics}
\usepackage[bookmarks=false]{hyperref}
\usepackage{graphicx}
\usepackage{subfig}
\usepackage{amsmath}
\usepackage{amssymb}

\title{A Novel 3D-UNet Deep Learning Framework Based on High-Dimensional Bilateral Grid for Edge Consistent Single Image Depth Estimation}

\name{Mansi Sharma$^{1}$, Abheesht Sharma$^{2}$, Kadvekar Rohit Tushar$^{3}$, Avinash Panneer$^{4}$}
\address{Department of Electrical Engineering, Indian Institute of Technology Madras, India$^{1,3,4}$ \\ 
Department of Computer Science \& Information Systems,
BITS Pilani, K K Birla Goa Campus$^{2}$}

\begin{document}
\ninept
\maketitle

\begin{abstract}
The task of predicting smooth and edge-consistent depth maps is notoriously difficult for single image depth estimation. This paper proposes a novel Bilateral Grid based 3D convolutional neural network, dubbed as 
3DBG-UNet, that parameterizes high dimensional feature space by encoding compact 3D bilateral grids with UNets and infers sharp geometric layout of the scene. Further, another novel 3DBGES-UNet model is introduced that integrate 3DBG-UNet for inferring an accurate depth map given a single color view. The 3DBGES-UNet concatenates 3DBG-UNet geometry map with the inception network edge accentuation map and a spatial object’s boundary map obtained by leveraging semantic segmentation and train the UNet model with ResNet backbone. Both models are designed with a particular attention to explicitly account for edges or minute details. Preserving sharp discontinuities at depth edges is critical for many applications such as realistic integration of virtual objects in AR
video or occlusion-aware view synthesis for 3D display applications.The proposed depth prediction network achieves state-of-the-art performance 
in both qualitative and quantitative evaluations on the challenging NYUv2-Depth data. The code and corresponding pre-trained weights will be made publicly available.

\end{abstract}

\begin{keywords}
Monocular depth estimation, computational photography, edge-aware image processing, convolutional neural network, deep learning, bilateral grid, 
VR/AR, 3D display.
\end{keywords}

\section{Introduction}
\label{sec:intro}
Monocular depth estimation is an ill-posed problem yet highly demanding for augmented or mixed reality, robotics and in general for 3D computer vision applications. With the advent of deep learning, especially convolutional neural networks, several methods have been proposed to solve this problem 
with desirable results. The algorithms either rely on supervised learning or self-supervised learning approaches \cite{RWRef1, RWRef2, RWRef3, RWRef4, RWRef5, RWRef6, RWRef7, RWRef8, RWRef9, RWRef10}.

\begin{figure}[t]
    \centering
    \subfloat{\includegraphics[width=85pt,height=65pt]{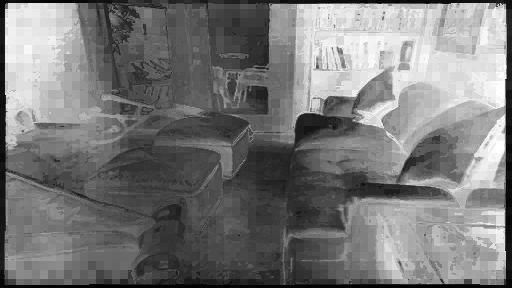}} 
    \hspace{0.1pt}
    \hspace{0.1pt}
    \subfloat{\includegraphics[width=85pt,height=65pt]{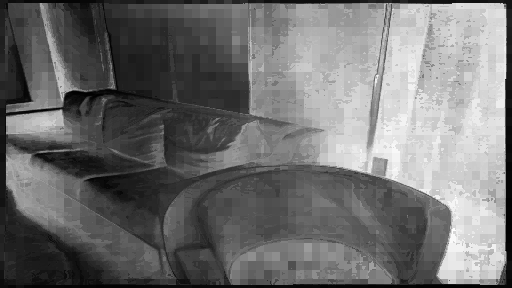}} 
    \hspace{0.1pt}
    \subfloat{\includegraphics[width=85pt,height=65pt]{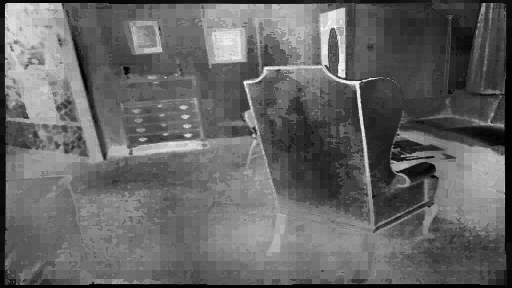}} 
    \hspace{0.1pt}
    \subfloat{\includegraphics[width=85pt,height=65pt]{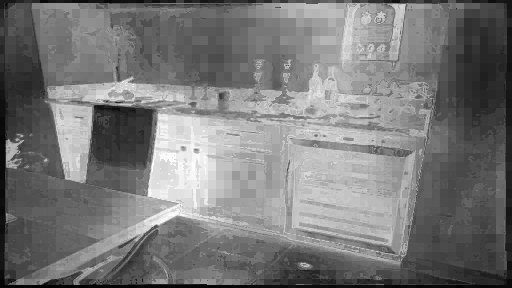}} 
\caption{Sharp geometric layout of the scene recovered using proposed 3DBG-UNet model.}
    \label{BLGridMaps}
\end{figure}

\begin{figure*}[t]
\centering
\includegraphics[width=430pt,height=150pt]{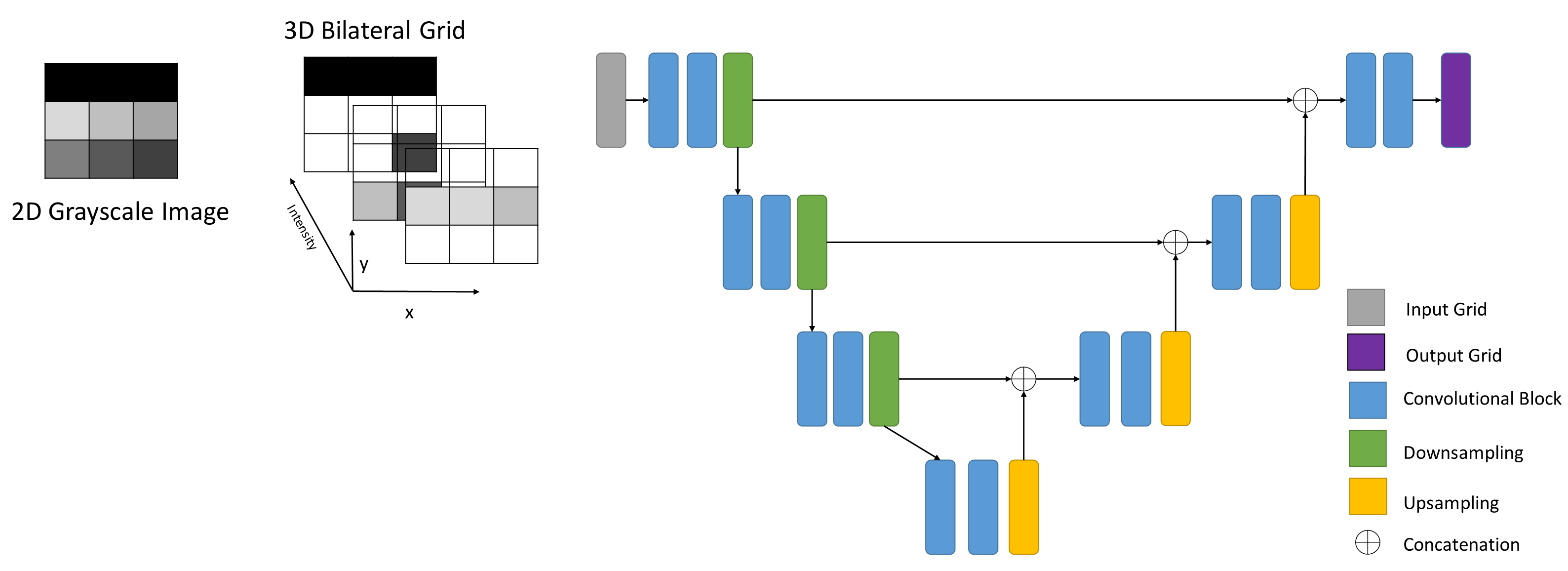}
\caption{An overview of proposed 3DBG-UNet architecture
for sharp geometric layout of the scene.
}
\label{fig:3DBL-UNet}
\end{figure*}

Despite recent advances in monocular depth estimation, preserving sharp discontinuities at depth edges or object's boundaries remain difficult to estimate correctly. Some methods incorporate additional information of sharp edges or occluding contours that constrain the depth predictions during learning. For example, Ramamonjisoa and Lepetit \cite{RWRef8} enforce depths and normals geometry consensus to learn the object's contours. Swami et al. \cite{RWRef9} added a fully differentiable variant of ordinal regression in depth estimates. Chen et al. \cite{RWRef8} progressively add finer structures while recovering coarser layout at multiple scales. In general, the major reasons of lack of precise edges or object's boundary details in predicting depth are many. First, the networks build on depth annotations of training images obtained with a stereo algorithm or a structured RGB-D camera are probable to be inaccurate around the object's boundaries. This is the case with NYUv2-Depth dataset \cite{DatasetRef1}. The 3D points around the edges or occluding contours may not be visible in all training views. The structured RGB-D cameras based on stereo technique, where one viewpoint is replaced with the known pattern suffer from the same problem. Secondly, the minute edges around objects and occluding regions may not influence the loss function during training unless explicitly handled with definite care. Most of the existing solutions for monocular depth estimation often produce blurry approximations of low resolution. While latest algorithms have shown steadily increasing performance, there are still major problems in both the quality and the resolution of estimating depth maps.
This paper addresses these critical problems in monocular depth estimation.
Our contributions are twofold:
\begin{itemize}

\item{We introduce a novel Bilateral Grid based 3D CNN model, dubbed as ‘3DBG-UNet’, that infers sharp geometric layout of the scene. Given a single color image, the model learns to generate a geometric map that captures minute details and object boundaries more faithfully. The key insight of the 
3DBG-UNet model is to naturally replace spatial convolution layers with 3D
bilateral grids and learn on high-dimensional image representation that 
enables localization and capture the scene context precisely.
} 

\item{We propose another new convolutional neural network, called 3DBGES-UNet, for computing detailed high resolution depth map given a single color view. This integrated model consists of a new building block (3DBG-UNet) that lifts the image onto a 3D bilateral grid in order to extract geometric information in an edge-preserving way. Subsequently, it combines the geometry map of 3DBG-UNet with the inception network edge accentuation map and a spatial object’s boundary map to refine the depth via a UNet architecture on the ResNet backbone. Numerical results indicate state-of-the-art performance on NYUv2-Depth data.
} 
\end{itemize}
The results demonstrate that integrating the geometry perspective of bilateral grids in our 3D UNet models augment the training and guiding a network
to learn better weights for edge consistent output with fewer parameters and less training iterations. Both proposed models are 
versatile and could be useful in providing complementary geometry-suggestions to any existing supervised or self-supervised depth prediction schemes,
not just our own.

\section{Related Work}
We briefly categorize the learning techniques according to the different generalizations of this work. We summarize algorithms for single-image depth estimation and also discussed models that learn general forms of high-dimensional data for vision and image processing applications.

\subsection{Monocular depth estimation}
We discuss the recent methods for monocular depth estimation using supervised and self-supervised learning technique. Godard et al. \cite{RWRef1} proposed a self-supervised monocular depth estimation method based on a minimum 
re-projection loss, an auto-masking loss, a full-resolution multi-scale sampling. The network deals with occlusions between frames in monocular video and flexible to train with monocular video data, stereo data, or mixed monocular and stereo data.

Tosi et al. \cite{RWRef2} leverage stereo matching in order to improve monocular depth estimation. The deep architecture, monoResMatch, proposed by them infers depth from a single image by synthesizing features from a virtual right viewpoint, horizontally aligned with the original view, and further performing stereo matching between the two cues. The network is trained end-to-end from scratch. 

Xu et al. \cite{RWRef3} integrate continuous Conditional Random Fields into an end-to-end deep architecture. This helps in fusing multi-scale information derived from different layers of a CNN. Their structured attention model automatically regulates the information transferred between corresponding features at different scales.

Watson et al. \cite{RWRef4} analysed ambiguous re-projection in depth-prediction from stereo-based self-supervision. They alleviate this effect by introducing complementary depth-suggestions, dubbed as Depth Hints. Such hints are computed from simple off-the-shelf stereo algorithms to enhance the photometric loss function and guiding the network to learn better weights. 

Hu et al. \cite{RWRef5} considered fusion strategy of extracting features at different scales and also minimize inference errors used in training by measuring loss functions. They reconstructed depth maps with finer resolution with small objects and object boundaries. Alhashim and Wonka \cite{RWRef6} transfer learning approach recovers high-resolution depth map that capture object boundaries. Their CNN model leverages standard encoder-decoder models alongside transfer learning to predict depth of a single RGB image. 

Ramamonjisoa and Lepetit \cite{RWRef7} SharpNet introduced geometry consensus in multi-task training to pay particular attention to occluding contours. Swami et al. \cite{RWRef9} proposed a fully differentiable variant of ordinal regression for depth estimation. They trained their network in end-to-end fashion. Chen et al. \cite{RWRef10} predict multi-scale depth maps by progressively adding finer structures at a specific scale while preserving the coarser layout. 
 
While the performance of monocular depth estimation methods has been increasing steadily, general problems with resolution, preserving accurate object's boundaries and quality of the predicted depth maps leave a lot of room for improvement. This paper utilizes the principle of bilateral grids in a clever way that separate pixels by their spatial and range dimension and allow edge-aware operations with simple 3D “fully convolutional network” for depth estimation at expected resolution. 

\begin{figure*}[t]
\centering
\includegraphics[width=430pt,height=180pt]{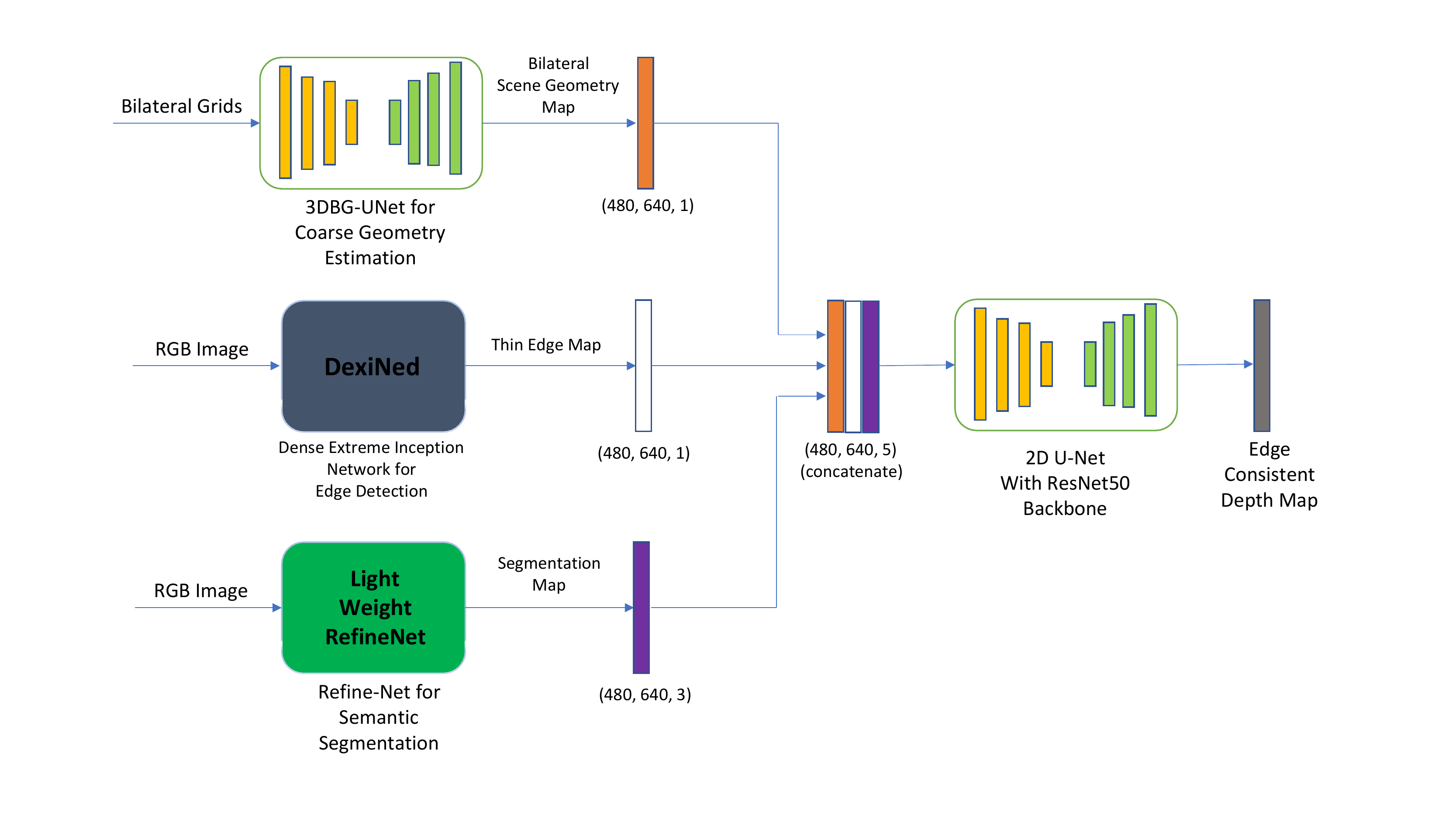}
\caption{An overview of the proposed integrated 3DBGES-UNet model for edge-consistent depth estimate from a single image.}
\label{fig:3DBLES-UNet}
\end{figure*}

\subsection{Neural networks for high-dimensional data processing} 
The CNNs designed for high-dimensional data representation made good progress in recent years. The diverse roles of machine learning algorithms where the coordinates of underlying data representation have a grid structure have been demonstrated for several computer vision and image/video processing applications. Bruna et al.~\cite{RWRef11} propose deep neural network on graphs following the convolutions on irregular grids. Their graph defines high-dimensional convolution using “locally” connected and pooling layers. This constraint the graph structure for adding new samples since the weights in the graph nodes determine a notion of locality. The relatively weak regularity assumption on their graph with fixed local receptive fields make construction complex to induce weight sharing across different locations, and thus reusability is a challenge with their costly spectral construction. 

Jaderberg et al. \cite{RWRef12} introduce a learnable module called the `Spatial Transformer', which explicitly allows the spatial data manipulation within the neural network. This spatial transformation capability enables to learn 
invariance for a given task in a parameter efficient manner. Ionescu et al. \cite{RWRef13} present a sound mathematical framework that generalizes adjoint matrix variations for backpropagation architectures. This enables global structured matrix computations, inclusion of structural layers such as normalized-cuts, higher-order pooling into deep computation architectures. Ben Graham \cite{RWRef14} extends 2D CNNs to sparse 3D CNNs. The experiments with CNNs on the 2D triangular-lattice and 3D tetrahedral-lattice are presented in \cite{RWRef14} for 3D object recognition and  space-time analysis of objects.

Jampani et al. \cite{RWRef15} suggests to replace the spatial convolution layers
of CNN architectures with bilateral filters, which have a varying spatial receptive field depending on the image content. This helps to overcome constraints caused by fixed local receptive fields, and enables 3D filtering with sparse samples handled by a permutohedral lattice data structure.

The idea proposed in this paper is absolutely different from one ``splat-blur-slice'' paradigm presented by Jampani et al. \cite{RWRef15} and other lattice based 3D CNN models \cite{RWRef11,RWRef12,RWRef13,RWRef14}. There is no processing performed inside the grids. Instead, we use the principle of bilateral grid data structure in respecting strong edges that come from the inclusion of the range in extra third dimension. This enables Euclidean edge-aware image manipulation meaningful with grid-based deep network layers in determining good localization and capturing scene context at the same time by learning from the data itself.
 
\section{Proposed CNN Architectures} 
Nonlinear filters have enabled a variety of image enhancement and editing or  manipulation techniques where image structure is taken into account. In particular, the bilateral filter is a nonlinear process that respects strong edges while smoothing an image \cite{Ref1}. Bilateral filtering has been used in a variety of contexts in computational photography or image processing applications such as tone mapping, inpainting, style transfer, denoising, relighting, and more applications. A common drawback of nonlinear filters such as bilateral filter is the computational complexity for processing high-definition content \cite{Ref1}.

Paris and Durand \cite{Paris} extend the fast bilateral filter presented by Durand and Dorsey \cite{Durand}. They recast the bilateral filtering as a higher-dimensional space linear convolution followed by trilinear interpolation and a division. This linear filtering process, further, speeds up the operation. Chen et al. \cite{Ref1} generalize the ideas behind Paris and Durand \cite{Paris} higher dimensional space into a new compact data structure, the \textit{bilateral grid}, that  enables a number of edge-aware image manipulations on high resolution images in real time. A bilateral grid is a 3D representation of a 2D image that separates the pixels not only by spatial position or coordinate, but also by respective intensity value or range coordinate. Chen et al. \cite{Ref1} defined bilateral grid as a three dimensional array, where the first two dimensions $(x, y)$ in spatial domain correspond to the 2D position in the image plane, while the third dimension $z$ corresponds to a reference range. The reference axis corresponds to the image intensity in most cases. Given a reference image $I$, Chen et al. \cite{Ref1} describes the image manipulation as a three step procedure. First, create a bilateral grid from an image. Then, perform processing inside the grid. Finally, extract a 2D value map by accessing the grid at $(x,y, I(x,y))$, typically performing data interpolation and slicing operation. Thus, processing on the grid between construction and slicing (symmetric operations) is what enables edge-aware operations in their approach. The required sampling rate is determined by the particular grid operation. In practice, most grid operations require only a coarse resolution. Usually, the number of grid cells is much smaller than the number of image pixels.

In our proposed framework, we do not perform the processing inside the grid. Instead, edge-aware properties of the bilateral grid are exploited in learning deep 3D-UNet model for edge consistent depth map prediction from the data itself. It is critical to note that in the spatial dimension of an image, although two pixels across an edge are close, but from the bilateral grid perspective, they are distant because their values differ widely in the range dimension of the data structure. This ensures that only a limited number of intensity values will get affected around edge pixels, while performing 3D convolution operations. Besides, modeling bilateral grids in CNN layers allow the network to  simultaneously localize in two-dimensional spatial domain while operating on one-dimensional range dimension to see the context.

We propose two architectures in the next section. The first 3DBG-UNet model is used for recovering sharp geometric layout of the scene. It operates taking bilateral grids as input and gives a bilateral grid as an output. The 3DBG-UNet architecture is inspired from previous 2D U-Net architecture of Ronneberger et al. \cite{Ref2}. All operations in our 3DBG-UNet are performed in three dimensional space. In particular, it is built on 3D convolutions, 3D max pooling, and 3D up-convolutional layers. Besides, we avoid bottlenecks in the network architecture for smooth processing of bilateral grids and use 3D batch normalization for faster convergence. 

The second one, dubbed as 3DBGES-UNet, is designed to recover detail high-resolution depth map. It combines geometric output of the 3DBG-UNet network with an edge and a semantic segmentation map to refine the final depth via a deep U-Net architecture. Both models are designed to learn from a single monocular image with a particular attention to faithfully account for fine details and prominent edges.

\subsection{Proposed 3DBG-UNet architecture for learning scene geometric layout}
\label{3DBG-UNetsection} 
The 3DBG-UNet architecture is illustrated in Fig.~\ref{fig:3DBL-UNet}. The first building block of 3DBG-UNet lifts the original image in a 3D bilateral grid. The high-dimensional image representation via bilateral grid is obtained by accumulating the value of each input pixel into the appropriate grid voxel.
Let a grayscale image $\textbf{I}$ be represented as $\textbf{I}(x,y) = \iota$, where $(x,y)$ encodes the 2D pixel position in the image plane and form the spatial domain. The $\iota$ denotes intensity value at the pixel position $(x,y)$. Let an image $\textbf{I}$ be normalized to $[0, 1]$. The corresponding bilateral grid $\Gamma_{BG}$ is constructed as follows:

\begin{itemize}
    \item{For all grid nodes $(a,b,c) \in \Gamma_{BG}$, initialize as,
    
    \begin{equation}
    \Gamma_{BG}(a,b,c)=(0,0)
    \end{equation}
    
    }
    
    \item{For each pixel position $(x,y)$:
    \begin{equation}
    \Gamma_{BG}([\frac{x}{sr_s}], [\frac{y}{sr_s}], [\frac{\iota}{sr_r}])
    \hspace{10pt} += (\iota, 1)
    \label{EQgrayBLG}
    \end{equation}
    
    } 
\end{itemize}
where, $[\cdot]$ is the closest-integer operator, $sr_s$ is sampling rate of the spatial axis, and $sr_r$ denotes the sampling rate of the range axis. The range axis is image intensity. The amount of smoothing is governed by $sr_s$, while the degree of edge preservation is controlled by $sr_r$.
\\
The bilateral grid representation (\ref{EQgrayBLG}) is constructed for a grayscale image. To handle color images, the higher dimensional $5D$ grids need to be created considering dimension for RGB color channels. Constructing and processing 5D grids in CNNs is practically difficult. In proposed formulation, we instead operate on three bilateral grids separately, one for each color channel. That is, we consider each color channel image as a grayscale image and apply (\ref{EQgrayBLG}) to it. Let the RGB image be represented as $I(x,y) = (\iota_r, \iota_g, \iota_b)$. We, therefore, represent the bilateral grid for each channel $c \in (r, g, b)$, as follows:

\begin{itemize}

    \item{For all grid nodes $(a,b,c) \in \Gamma_{BG_c}$, initialize as,
    \begin{equation}
    \Gamma_{BG_c}(a,b,c)=(0,0)
    \end{equation}
    }
    
    \item{For each pixel position $(x_c,y_c)$:
    \begin{equation}
    \Gamma_{BG_c}([\frac{x_c}{sr_s}], [\frac{y_c}{sr_s}], [\frac{\iota_c}{sr_r}])
    \hspace{10pt} += (\iota_c, 1)
    \label{EQcolorBLG}
    \end{equation}
    } 
\end{itemize}

After lifting the original image in high dimensional space, the bilateral grid is processed along the 3D contracting path and a 3D expanding path. On the left side, the architecture consists of a 3D contracting path. A 3D expanding path is depicted on the right side. The typical architecture along the 3D contracting path, \textit{i.e.} the encoder of 3DBG-UNet, is composed of two 3D sub-blocks corresponding to each block. Each sub-block consists of a 3D convolutional layer with a kernel size of $5$, stride of $1$ and padding of size $2$. This is followed by a rectified linear unit (ReLU) which act as the activation function and a 3D batch-normalization layer \cite{NewRef1}. Each encoder block is followed by 3D max-pooling layer with kernel size $2$. The 3D max-pooling layer is used to perform downsampling between blocks of the encoder.

Along the expanding path of the network, each block contains repeated two sub-blocks similar as in the encoder. Every step in the expanding path consists of an upsampling of the feature map, a concatenation with the output of the corresponding encoder layer from the contracting path and the previous decoder layer. For upsampling between decoder blocks, a transposed 3D convolutional layer is employed with a kernel size of 4, stride of 2 and padding of size 1.
The feature channels in upsampling part in 3DBG-UNet allows to propagate scene context and layout details to higher resolution layers. The final layer in our proposed 3DBG-UNet architecture consists of sigmoid activation, which restricts the pixel values of the Bilateral geometry grid to the range $[0,1]$. 

The Mean Squared Error or MSE is the default loss function used in the proposed model. The MSE loss is defined as follows:
\begin{equation}
    MSEloss = \frac{1}{n} \sum_{y_i\in |n|} (y_{i}^{*} - y_{i})^{2}
\end{equation}
where $y^{*}$ is the predicted depth map and $y$ is the ground truth depth map, and $n$ is the number of pixels in the batch-size training images. The 3DBG-UNet model is trained for 150 epochs on NYU-Depth v2 data using around 20K samples for training and 694 samples for testing. 

The 3DBG-UNet architecture is easily regulated for RGB images by modifying
the number of input channels from $1$ to $3$. We make some modifications
to extend all operations in 3D along the contracting and expanding path 
of 3DBG-UNet. First, the image's resolution can be quite large, and adding a third dimension can cause a increase in grid size. Therefore, we decrease the spatial resolution of the bilateral grids by $2$ or $4$.
Usually, the range of the pixel values in images be $0-255$. However, since bilateral grids will have a third dimension representing the intensity or range axis, the size will be large if we take the range of the pixel values to be $0-255$. Thus, we restrict the pixel values to lie in a smaller range. If the range is $0-31$, the intensity dimension is $32$, i.e., we choose a sampling rate for the reference axis to be $8$. If the range is $0-63$, the intensity dimension is $64$, respectively, i.e., the sampling rate for the reference axis is $4$. 

\subsection{Proposed 3DBGES-UNet for monocular depth estimation}
We proposed a new 3DBGES-UNet architecture for monocular depth estimation based on 3DBG-UNet model. The coarse geometric layout of the scene recovered by 
3DBG-UNet model is further processed for estimating refined depth map from single monocular image. The geometry map of the scene computed from 3DBG-UNet, following the procedure in section \ref{3DBG-UNetsection}, is highly pixelated, but well highlights the edges and preserve object's boundaries (Fig.~\ref{BLGridMaps}). This is due to the fact that we reduce the spatial dimensions of the image, and keep only $0-31$ or $0-63$ range for the intensity or range dimension for fast computation. However, this limits to encode the depth variance in the scene properly. We refine the geometry output in our 3DBGES-UNet architecture. The integrated model is illustrated in Fig.~\ref{fig:3DBLES-UNet}.

The 3DBG-UNet geometry output is converted into the grayscale image.
The geometric information recovered has pronounced edges or sharp details (see Fig.~\ref{BLGridMaps}). To recover high-resolution depth map, the 3DBGES-UNet concatenate 3DBG-UNet geometry map with the inception network edge accentuation map and a spatial semantic segmentation map, both extracted from the RGB color image.

We adopt a Light-Weight RefineNet \cite{Ref4} for semantic image segmentation. ResNet-152 is considered as the backbone in \cite{Ref4}. The semantic segmentation information distinguishes boundary of the object 
in an image and helps in refining pixelated output of 3DBG-UNet. Further, we adopt a DexiNed model \cite{Ref5} for accounting thin edge-maps that are plausible for human eyes, but not discernible in 3DBG-UNet course geometric layout.

We leverage information from the 3DBG-UNet geometry map, semantic segmentation map and the edge map in our proposed integrated 3DBGES-UNet model. The shapes of bilateral-grid geometry map, the segmentation map and the edge map are $(480,640,1)$, $(480,640,3)$ and $(480,640,1)$ respectively. The three maps in parallel are concatenated along the channel axis to form a tensor of the shape $(480,640,5)$. Further, it is fed into a 2D-UNet model with a ResNet-50 backbone \cite{Ref3} and trained end-to-end. The 3DBGES-UNet model is trained for 150 epochs on NYU-Depth v2 data considering 20K samples for training and 694 samples for testing.

\begin{figure*}[t]
    \centering
    \subfloat{\includegraphics[width=85pt,height=65pt]{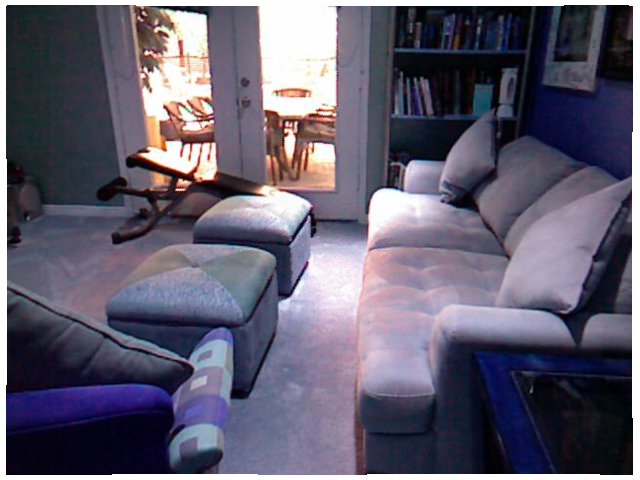}} 
    \subfloat{\includegraphics[width=85pt,height=65pt]{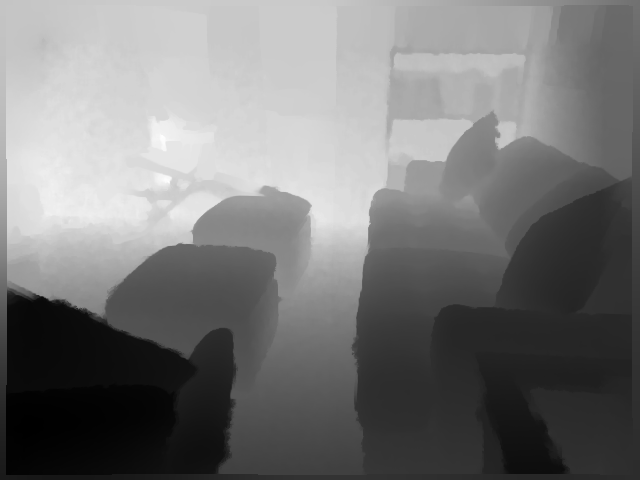}} 
    \subfloat{\includegraphics[width=85pt,height=65pt]{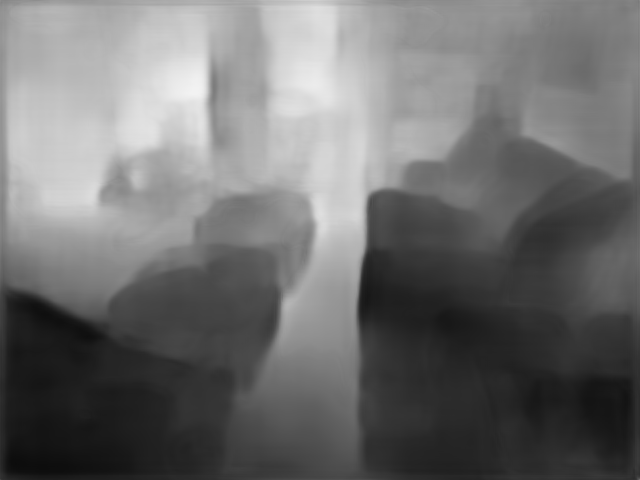}} 
    \subfloat{\includegraphics[width=85pt,height=65pt]{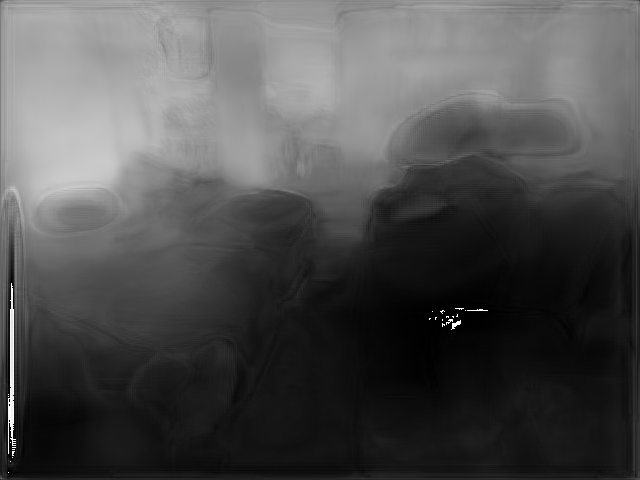}}
    \subfloat{\includegraphics[width=85pt,height=65pt]{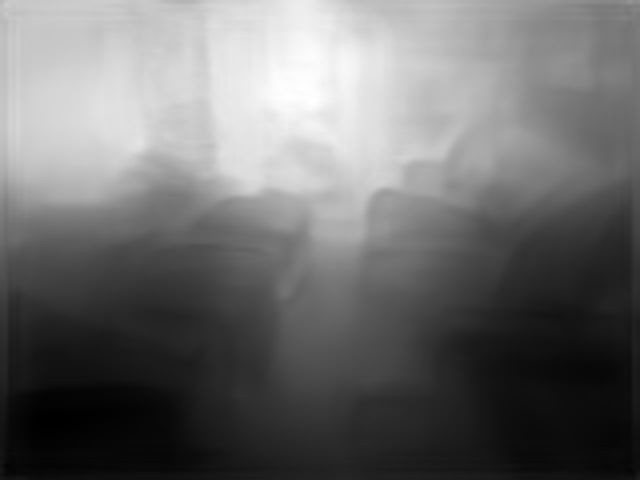}}
    \subfloat{\includegraphics[width=85pt,height=65pt]{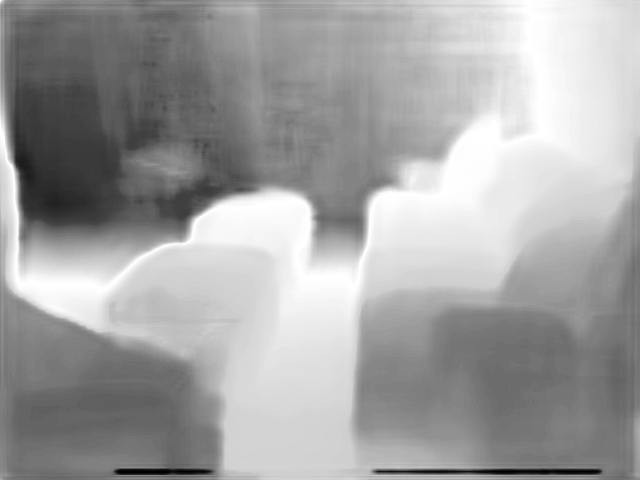}} 
    \\
    \subfloat{\includegraphics[width=85pt,height=65pt]{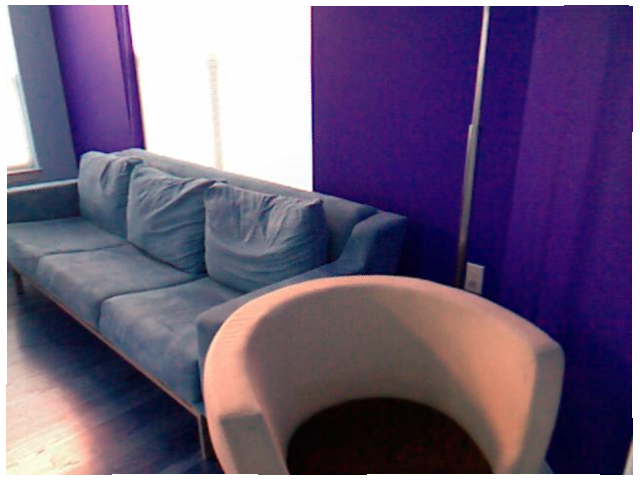}} 
    \subfloat{\includegraphics[width=85pt,height=65pt]{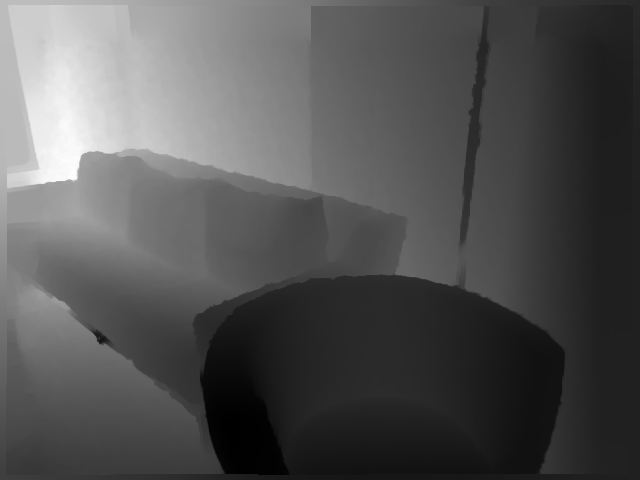}} 
    \subfloat{\includegraphics[width=85pt,height=65pt]{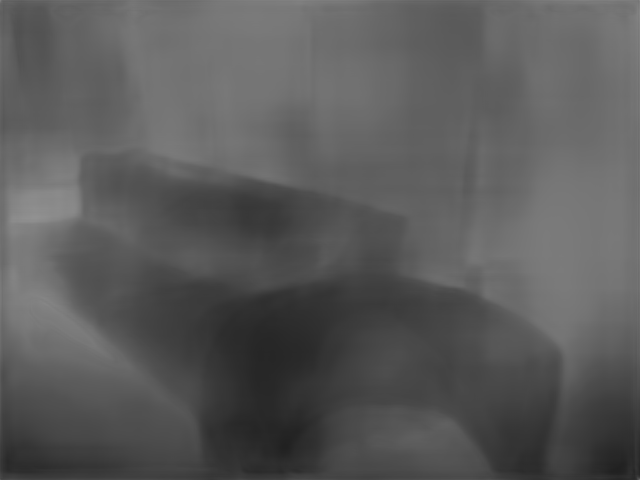}} 
    \subfloat{\includegraphics[width=85pt,height=65pt]{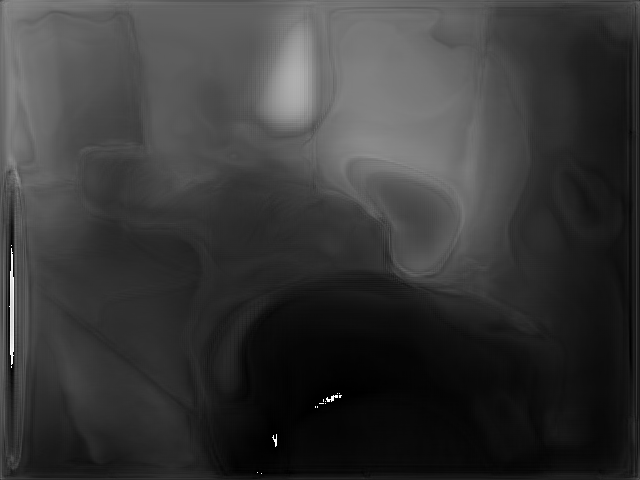}}
    \subfloat{\includegraphics[width=85pt,height=65pt]{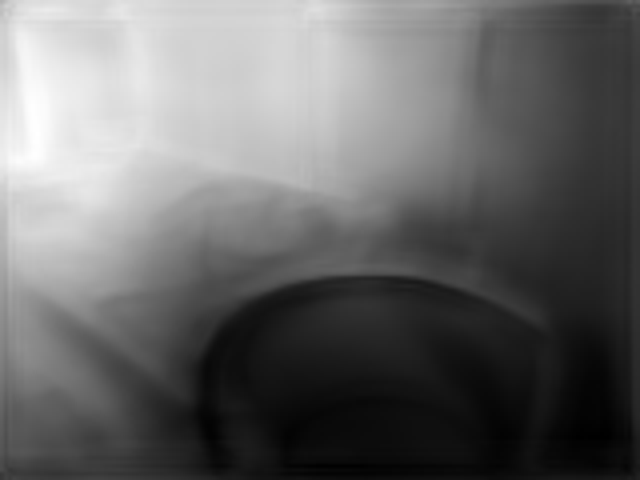}}
    \subfloat{\includegraphics[width=85pt,height=65pt]{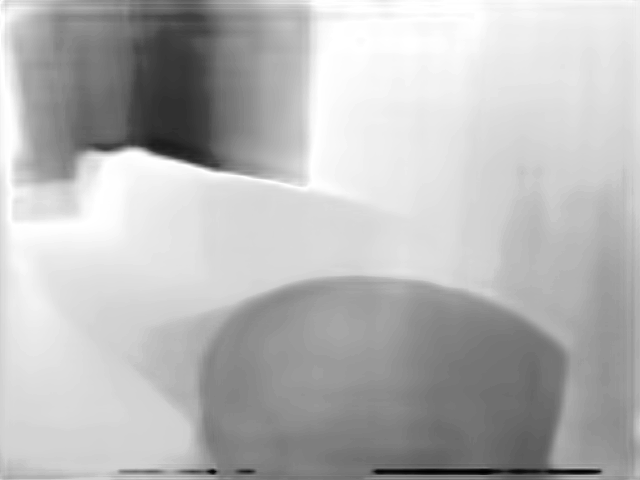}} 
\\
    \subfloat{\includegraphics[width=85pt,height=65pt]{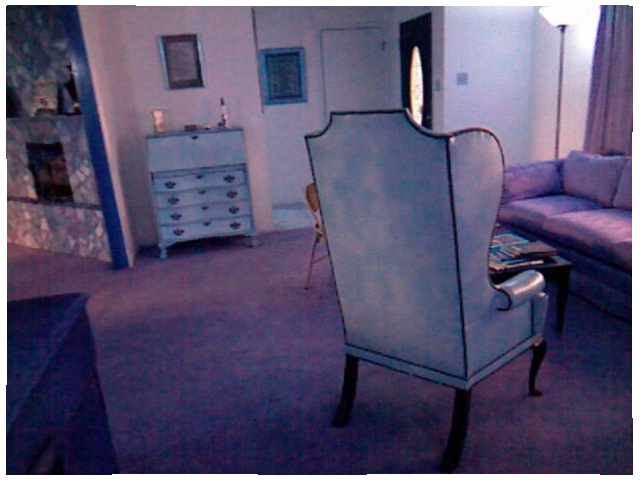}} 
    \subfloat{\includegraphics[width=85pt,height=65pt]{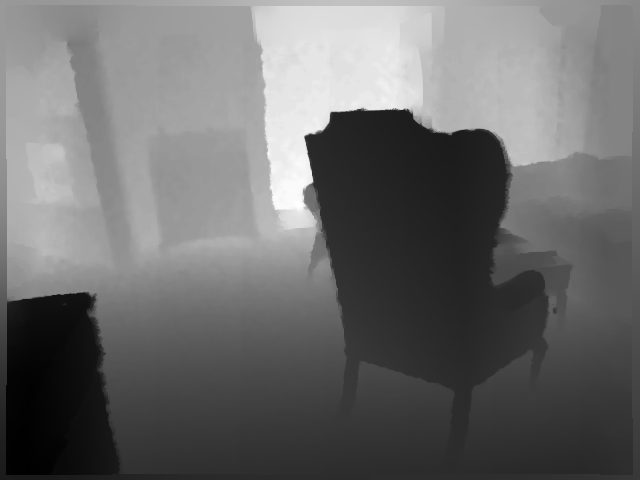}} 
    \subfloat{\includegraphics[width=85pt,height=65pt]{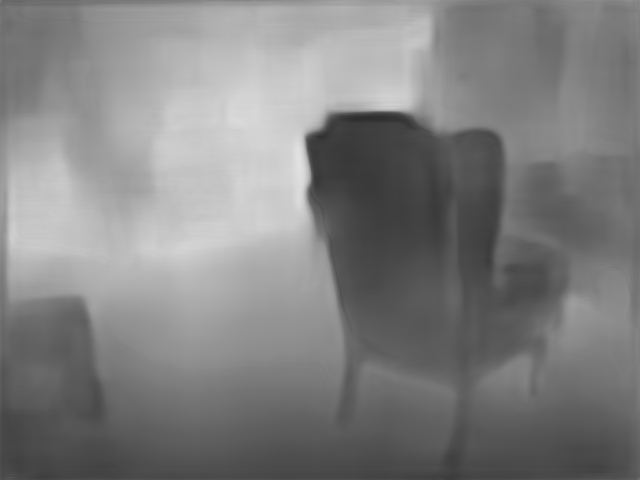}} 
    \subfloat{\includegraphics[width=85pt,height=65pt]{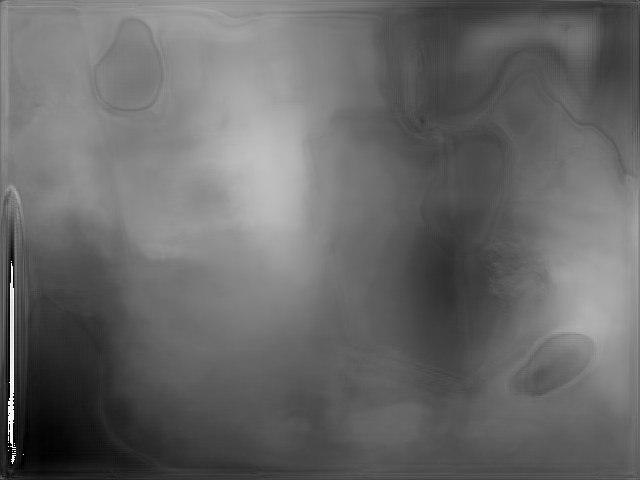}}
    \subfloat{\includegraphics[width=85pt,height=65pt]{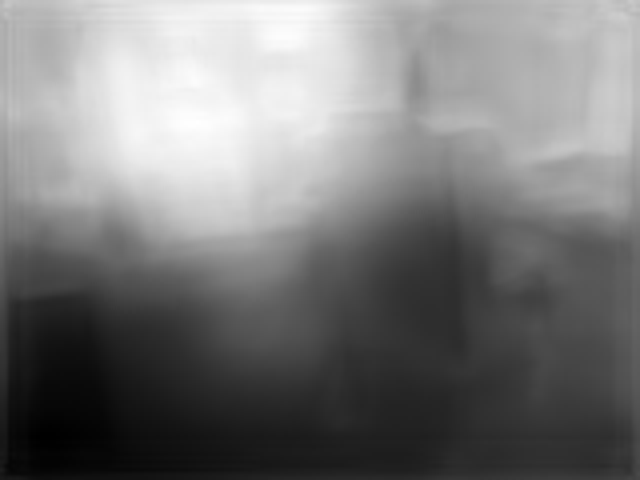}}
    \subfloat{\includegraphics[width=85pt,height=65pt]{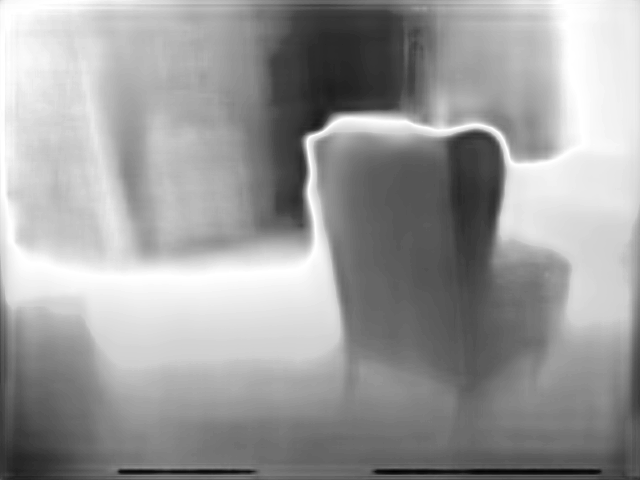}}
\end{figure*}

\begin{figure*}[t]
\centering
    \vspace{-0.5cm}
    \subfloat[Input]{\includegraphics[width=85pt,height=65pt]{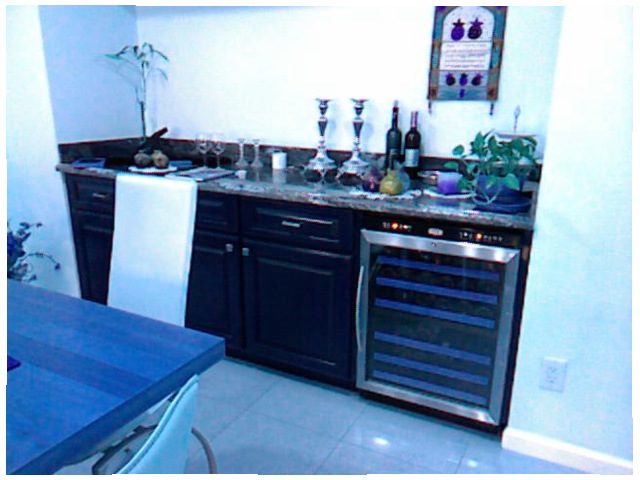}\label{fig:f1}}
    \subfloat[GT]{\includegraphics[width=85pt,height=65pt]{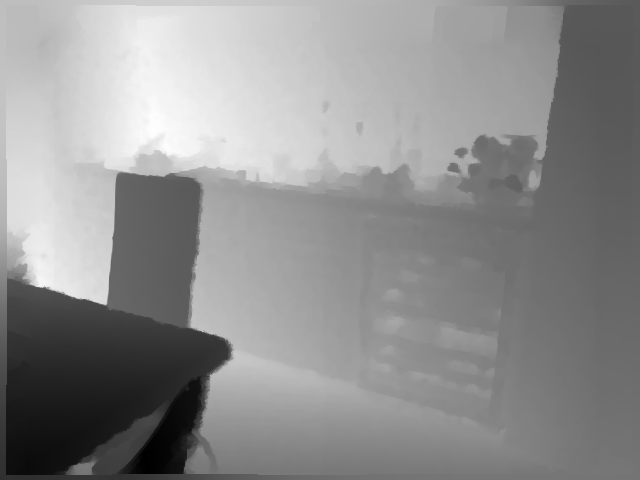}\label{fig:f2}}
    \subfloat[3DBGES-UNet (Ours)]{\includegraphics[width=85pt,height=65pt]{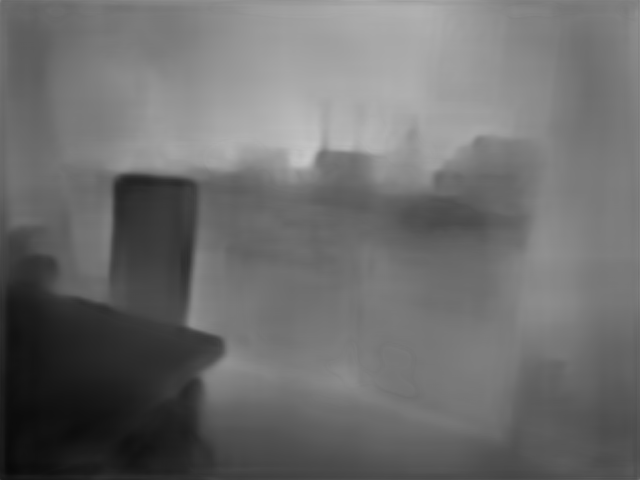}\label{fig:f3}}
    \subfloat[Serial U-Net]{\includegraphics[width=85pt,height=65pt]{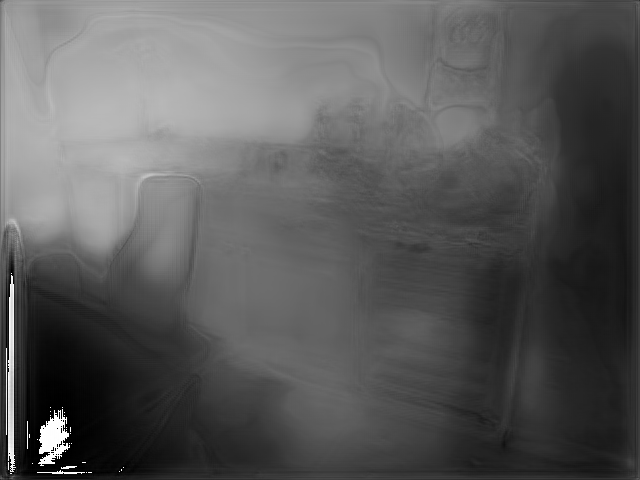}\label{fig:f4}}
    \subfloat[MSDN]{\includegraphics[width=85pt,height=65pt]{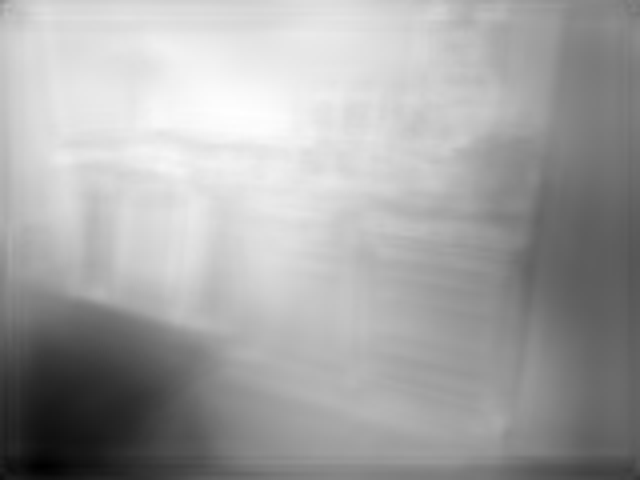}\label{fig:f5}}
  \subfloat[SharpNet]{\includegraphics[width=85pt,height=65pt]{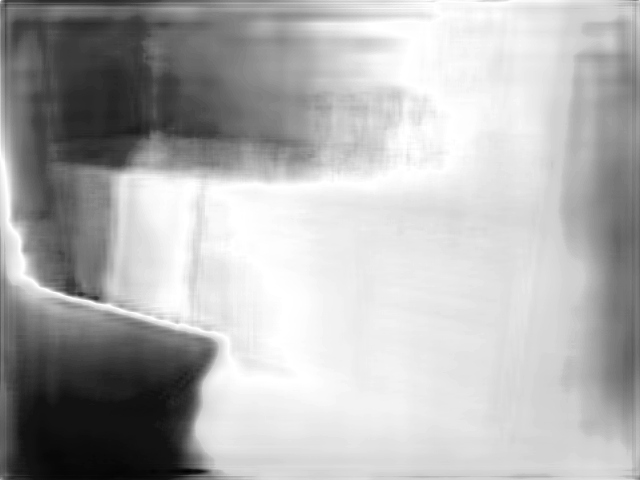}\label{fig:f6}}
   \caption{\textbf{Visual comparison of estimated depth maps.}}
    \label{CompareDepthFigure}
\end{figure*}

\section{Experiments and Results}
In this section, we describe experimental results and perform comparative analysis of proposed 3DBGES-UNet model with the state-of-the-art CNN algorithms for monocular depth estimation. We evaluated different methods on benchmark NYU-Depth V2 dataset in order to compare the robustness of our models.

\subsection{Implementation Details}
We implemented proposed 3DBG-UNet and 3DBGES-UNet models in PyTorch.
Both training and evaluation are performed on a single high-end HP OMEN X 15-DG0018TX Gaming laptop with 9th Gen i7-9750H, 16 GB RAM, RTX 2080 8 GB Graphics and Windows 10 operating system. Training took around 8 hours for 
the 3DBG-UNet and 11 hours for the 3DBGES-UNet network. Test prediction takes 
0.61 seconds per image with 3DBG-UNet model and 0.18 seconds per image with 
3DBGES-UNet model.

\subsubsection{NYU-Depth V2 data}
NYU-Depth V2 dataset is comprised of a variety of indoor scenes, captured by Microsoft Kinect as RGB-D video sequences \cite{DatasetRef1}. The color images and depth maps are provided at a resolution of $640 \times 480$. The training set has 120K samples \cite{CompareRef2}. We train our models on a 20K subset with filled in depth values using the colorization scheme of \cite{Inpaintref1}. We validated on 694 image pairs from the NYU-Depth v2 test data.

\subsubsection{Evaluation \& Comparative Analysis}
We quantitatively compare our 3DBGES-UNet model with CNN algorithms Cantrell et al. \cite{Ref3} (Serial U-Net), Ramamonjisoa et al. \cite{RWRef8} (SharpNet), Eigen et al. \cite{CompareRef2} (MSDN) in Table~\ref{TableI}. We evaluate these methods using most common error metrics from prior works \cite{Ref3,RWRef8,CompareRef2}. The error metrics are defined as:

\begin{itemize}
    \item{root mean squared error (RMSE): 
    
    \begin{equation}
    \sqrt{\frac{1}{n} \sum_{p}^{n} (y_p - \hat{y}_p)^{2}}
    \end{equation}
    
    } 
    \item{average ($log_{10}$) error: 
    
    \begin{equation}
    \frac{1}{n} \sum_{p}^{n} |log_{10} (y_p) - log_{10} (\hat{y}_p)|
    \end{equation}
    
    } 

\end{itemize}
where, $y_p$ is a pixel in depth image $y$, $\hat{y}_p$ is a pixel in the predicted depth image $\hat{y}$, and $n$ is the total number of pixels for 
each depth image. Note that numerical results in Table~\ref{TableI} might vary from the original papers because we evaluated these methods with the same
code and  used the authors provided pre-trained models to compute their depth map predictions.

In Table~\ref{Table3}, we compare RMSE values of several state-of-the-art algorithms on NYU-Depth V2 data given at the website \cite{SOTAresults}. We choice to compare models not using extra training data. The comparison is performed with Zhu et al. \cite{NewMethod1} (SOM), Xu et al. \cite{NewMethod2}
DeepLabV3+ \cite{NewMethod3}, Multi-Task Light-Weight-RefineNet	\cite{NewMethod4}, RelativeDepth \cite{NewMethod5}, SC-SfMLearner-ResNet18	\cite{NewMethod6}, SDC-Depth \cite{NewMethod7}, ACAN \cite{NewMethod8}, DORN \cite{NewMethod9}, and SENet-154 \cite{NewMethod10} depth estimation methods.
The results in Table~\ref{TableI} and Table~\ref{TableII} are quite impressive and on-par with state-of-the-art methods.
\\
\textbf{Qualitative Analysis:} 
We perform qualitative analysis of different methods on the NYU-Depth V2 test data. The perception-based qualitative metric and depth edge reliability metric
are computed, as defined by \cite{RWRef6}. We computed the mean structural similarity score (mSSIM) that measures the similarity of resulting depth maps in the image space. The mSSIM scores are computed considering gray scale visualization of the ground truth as a reference and estimated depth map as a predicted image. Mathematically, the mSSIM quality score is computed as:
\begin{equation}
      \frac{1}{N} \sum_{i}^{T} SSIM(y_i, \hat{y}_i)
\end{equation}
We assess the mean structural similarity term over the entire NYU-Depth V2 test data.

The second measure, depth edge reliability metric (DERM), analyse edges in the predicted depth maps and ground truth. The gradient magnitude image of both the ground truth and the predicted depth image are determined for each sample.
The standard Sobel gradient operator is considered to generate gradient of the depth image \cite{QA1}. Further, the gradient magnitude image is threshold
and the $F$ score (also called $F_1$ score) is determined to measure the model accuracy \cite{QA2}. We choose threshold values greater than 0.5. The $F$ (or $F_1$) scores given in Table~\ref{TableI} and Table~\ref{TableII} are averaged across the test set.

The visual comparison results are shown in Fig.~\ref{CompareDepthFigure}. Our proposed 3DBGES-UNet model produces higher quality depth maps compare with those generated by other CNN algorithms. The edges are better predicted and matched those of the ground truth with fewer artifacts. 

\subsection{Relevance of 3DBG-UNet model} 
\begin{figure*}[t]
\centering
    \vspace{-0.6cm}
    \subfloat[Input]{\includegraphics[width=85pt,height=65pt]{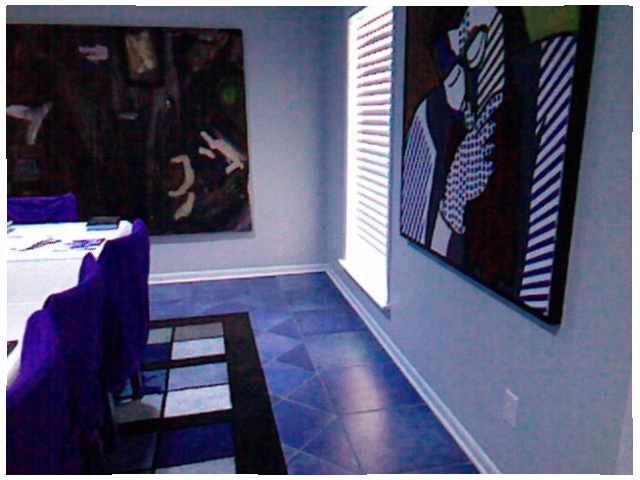}}
    \subfloat[GT]{\includegraphics[width=85pt,height=65pt]{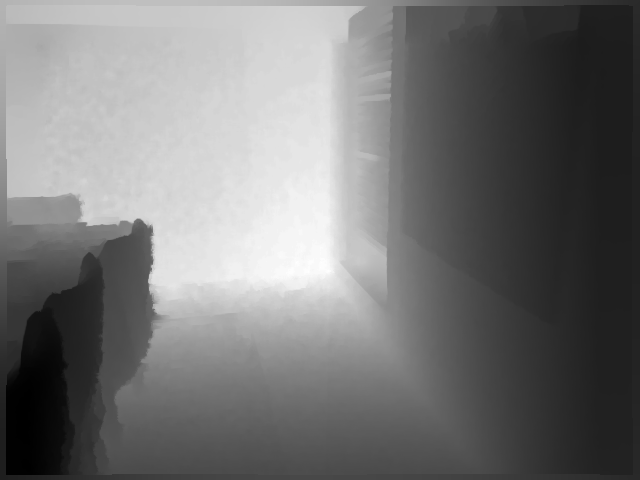}}
    \subfloat[3DBGES-UNet (Ours)]{\includegraphics[width=85pt,height=65pt]{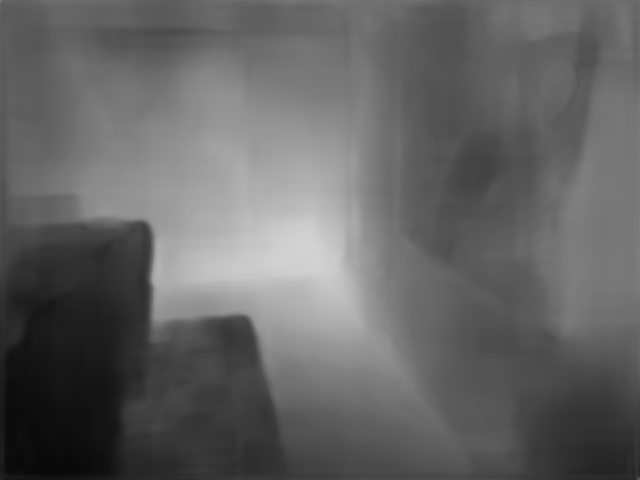}}
    \subfloat[RGB+Seg+Edge]{\includegraphics[width=85pt,height=65pt]{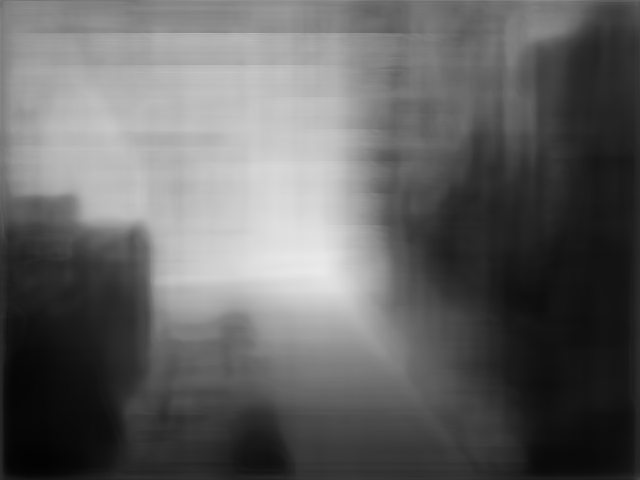}}
    \subfloat[RGB+Seg]{\includegraphics[width=85pt,height=65pt]{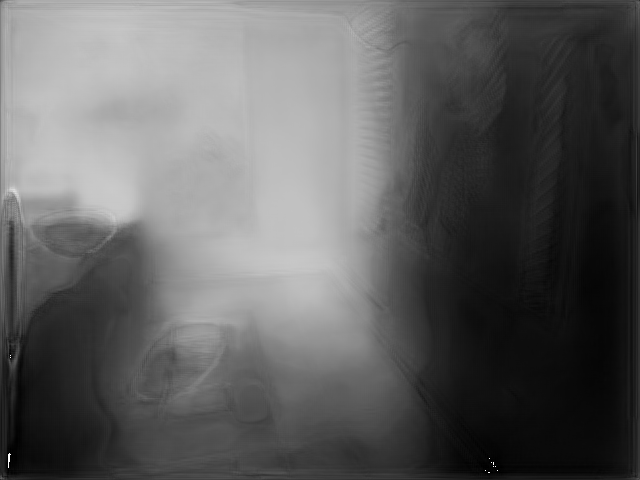}}
    \subfloat[RGB+Edge]{\includegraphics[width=85pt,height=65pt]{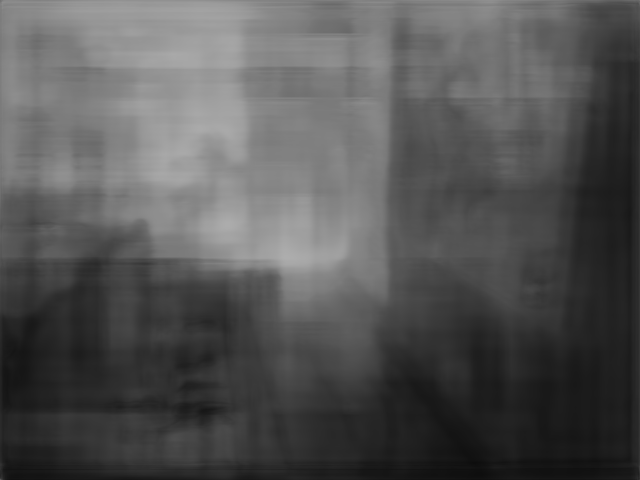}}
    \caption{\textbf{Visual comparison of 3DBGES-UNet against models with different combination of inputs.}}
   \label{CompareSubmodels}
\end{figure*}

We performed a critical experiment to assert the significance of our 3DBG-UNet model in the integrated 3DBGES-UNet model. We consider different combinations of inputs and trained three models, called as ``RGB+Seg+Edge'', ``RGB+Seg'' and  ``RGB+Edge'' respectively. In ``RGB+Seg+Edge'' model, we concatenate the RGB image (instead of considering 3DBG-UNet CNN model geometry map), the segmentation map (Seg), and the edge map (Edge). The RGB+Seg+Edge inputs are fed to the 2D-UNet stage of the architecture (See Fig.~\ref{fig:3DBLES-UNet}). In ``RGB+Seg'' model, we concatenate the RGB image and the segmentation map and fed to the 2D-UNet. Likewise, in ``RGB+Edge'' model, the RGB image and the edge map inputs are fed to the 2D-UNet stage respectively. All models are trained for 150 epochs. The output of three different models is compared with our 3DBGES-UNet model. The metric values in Table~\ref{TableII} validate that the depth quality decreases with a different combination of input in the three models. Thus, the numerical results in Table~\ref{TableII} and depth maps depicted in Fig.~\ref{CompareSubmodels} established that 3DBG-UNet CNN geometry map is essential in our 3DBGES-UNet model for inferring high quality detailed depth map with sharp edges and object's boundaries.

\begin{table*}
    \centering
     \caption{Comparison of different methods on the NYU-Depth V2 data. Higher values indicate better quality for mSSIM and DERM measures while lower values are better for RMSE and $log_{10}$.
     }
    \begin{tabular}{|c|c|c|c|c|c|c|c|c|c|}
    \hline
        & Serial U-Net \cite{Ref3}  & SharpNet \cite{RWRef8} & MSDN \cite{CompareRef2} & Ours  & Ours & Ours & Ours & Ours  \\
        \hline
         &  &  &   & epochs 50 & epochs 75 & epochs 100 & epochs 125 & epochs 150 \\
        \hline
         RMSE $\downarrow$  & 0.2089 & 0.4654 &   0.2071 & 0.1907 & 0.1929 & 0.1885 & 0.1855 & 0.1857  \\ 
         \hline
        $log_{10}$ $\downarrow$ & 0.2442 &  0.3926 & 0.2093 & 0.2061 & 0.2083 & 0.2112 & 0.2069 & 0.2070\\ 
         \hline
        mSSIM $\uparrow$ & 0.7482 & 0.5440 & 0.8042 & 0.8980 & 0.9010 & 0.9072 & 0.9044 & 0.9145
         \\
         \hline
        DERM $\uparrow$ &  0.3608 &  0.4685 & 0.3334  &  0.6266 & 0.6256 & 0.6325 & 0.6446  & 0.6523 \\
        \hline
        
        \end{tabular}
   
    \label{TableI}
    
\end{table*}

\begin{table*}
    \centering
     \caption{Relevance of 3DBG-UNet model in 3DBGES-UNet model. Here higher values indicate better quality for mSSIM and DERM measures, while lower values are better for RMSE and $log_{10}$.}
     
         \begin{tabular}{|c|c|c|c|c|c|c|c|c|c|}
         \hline
         & RGB+Seg+Edge  & RGB+Seg & RGB+Edge & Ours  \\
         \hline
         RMSE $\downarrow$  & 0.2078 & 0.2089 & 0.2321 &  0.1857 \\ 
         \hline
         $log_{10}$ $\downarrow$ & 0.2456 & 0.2442 &  0.2745 &  0.2070\\ 
         \hline
         mSSIM $\uparrow$ & 0.7758 & 0.7482 & 0.7565 &  0.9145\\
         \hline
         DERM $\uparrow$ &  0.4222 & 0.3608 &  0.4007 &  0.6523\\
         \hline
        
        \end{tabular}
   
    \label{TableII}
   
\end{table*}

\begin{table}
    \centering
     \caption{Comparison of RMSE values of proposed depth estimation method with state-of-the-art models on NYU-Depth V2 data \cite{SOTAresults}.}
     
         \begin{tabular}{|c|c|c|c|c|c|c|c|c|c|}
    \hline
       Methods & RMSE   \\
        \hline
        Zhu et al. \cite{NewMethod1} (SOM)   & 0.604\\ 
         \hline
         Xu et al. \cite{NewMethod2} & 	0.586 \\ 
         \hline
         DeepLabV3+ \cite{NewMethod3} & 0.575 \\ 
         \hline
         Multi-Task Light-Weight-RefineNet	\cite{NewMethod4} & 0.565 \\
         \hline
         RelativeDepth	\cite{NewMethod5} & 0.538 \\
         \hline
         SC-SfMLearner-ResNet18	\cite{NewMethod6} & 0.536\\
         \hline
         SDC-Depth \cite{NewMethod7} &	0.497 \\
         \hline
         ACAN \cite{NewMethod8} &	0.496 \\
        \hline
         DORN \cite{NewMethod9} & 0.509 \\
        \hline
        SENet-154 \cite{NewMethod10} & 0.530	\\
        \hline
         Our (3DBGES-UNet) &  0.1857 \\
        \hline
        
        \end{tabular}
   
    \label{Table3}
   
\end{table}

\begin{figure*}
    \centering
    \subfloat{\includegraphics[width=85pt,height=65pt]{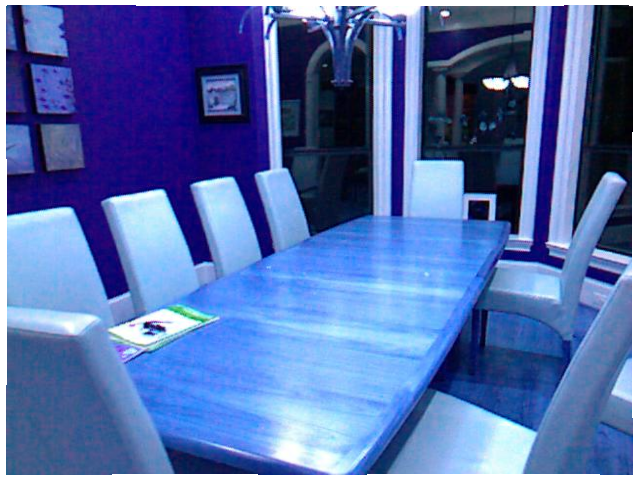}}
    \subfloat{\includegraphics[width=85pt,height=65pt]{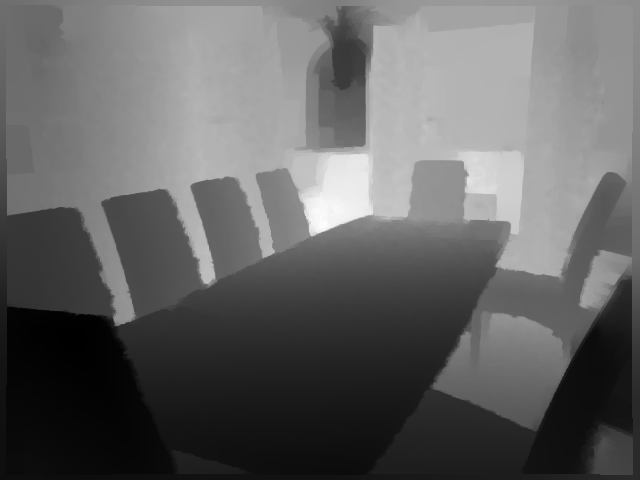}}
    \subfloat{\includegraphics[width=85pt,height=65pt]{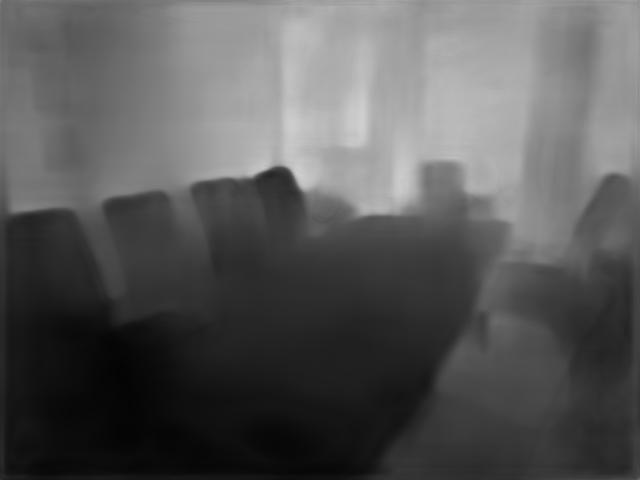}}
    \subfloat{\includegraphics[width=85pt,height=65pt]{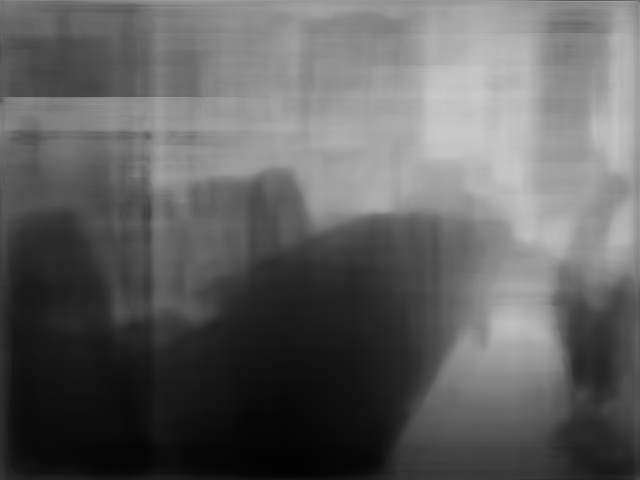}}
    \subfloat{\includegraphics[width=85pt,height=65pt]{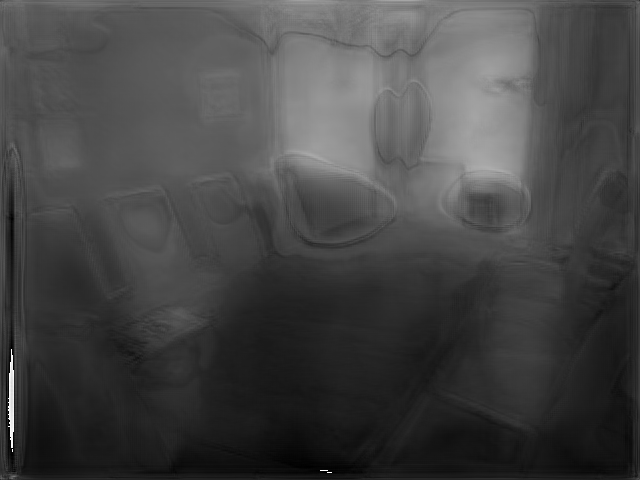}}
    \subfloat{\includegraphics[width=85pt,height=65pt]{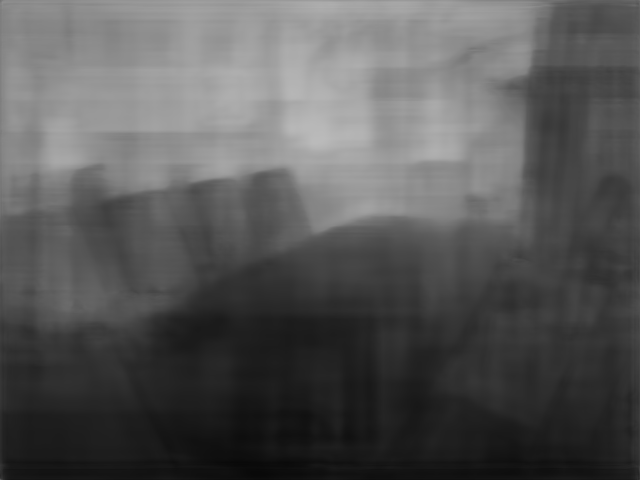}}
    \end{figure*}




\section{Conclusions}
We proposed novel deep neural network architectures, 3DBG-UNet and 3DBGES-UNet, based on high dimensional bilateral grid space for detailed high-resolution depth estimation from a single RGB image. Integrating geometry perspective of bilateral grids with learning strategies enable faithful capturing of strong edges and object boundaries in a scene. Evaluation on challenging benchmark NYUv2-Depth dataset demonstrate that our proposed model achieves state-of-the-art performance, both quantitatively and qualitatively. Our proposed self-supervised 3DBG-UNet is a versatile model that can provide complementary geometry-suggestions to any existing supervised or self-supervised depth prediction schemes, not just our own, to augment the training and guiding a network to learn better weights for edge consistent output. 

Following our simple 3DBG-UNet architecture, one avenue for future work is to combine with other best practices. We plan to substitute proposed bilateral grid encoder-decoder with a more compact one with parallel architectures, associated design choices with an effective loss functions and train with monocular video data, stereo data, or mixed monocular and stereo data, for superior predictions. 
Another future direction of research is to consider how to efficiently handle higher dimensional 6D bilateral grids in proposed learning approach for depth extraction from video. We shall look for another possibility in coupling dimensionality reduction techniques and plugging proxy-supervision in Bilateral CNNs at training time that could lead to more efficient networks. We will further explore the possibilities of deploying proposed ideas on interesting applications such as image manipulation or editing, image-to-image transfer, 3D view rendering, contrast enhancement, refocusing, object recognition, realistic integration of virtual objects in augmented reality.

\end{document}